\documentclass{article}

\usepackage{microtype}
\usepackage{graphicx}
\usepackage{subcaption}
\usepackage{booktabs} 

\usepackage{hyperref}

\usepackage[accepted]{icml2018}

\icmltitlerunning{Memorize or generalize? Searching for a compositional RNN in a haystack}

\usepackage[T1]{fontenc}
\usepackage[utf8]{inputenc}

\newcommand{\seg}[1]{\emph{#1}}
\newcommand{\bas}[1]{\emph{#1}}
\newcommand{\io}[1]{\guillemotleft\texttt{#1}\guillemotright}

\begin{document}

\twocolumn[
\icmltitle{Memorize or generalize? \\ Searching for a compositional RNN in a haystack}



\icmlsetsymbol{equal}{*}

\begin{icmlauthorlist}
\icmlauthor{Adam Li\v{s}ka}{fair,cncs}
\icmlauthor{Germ\'{a}n Kruszewski}{fair}
\icmlauthor{Marco Baroni}{fair}
\end{icmlauthorlist}

\icmlaffiliation{fair}{Facebook Artificial Intelligence Research}
\icmlaffiliation{cncs}{Center for Neuroscience and Cognitive Systems @ UniTn, 
Istituto Italiano di Tecnologia, Rovereto, Italy}

\icmlcorrespondingauthor{Adam Liska}{adam.liska@gmail.com}

\icmlkeywords{Machine Learning, ICML}

\vskip 0.3in
]



\printAffiliationsAndNotice{}  

\begin{abstract}
  Neural networks are very powerful learning systems, but they do not readily
    generalize from one task to the other. This is partly due to the fact that
    they do not learn in a \emph{compositional} way, that is, by discovering
    skills that are shared by different tasks, and recombining them to solve
    new problems. In this paper, we explore the compositional generalization
    capabilities of recurrent neural networks (RNNs). We first propose the
    \emph{lookup table composition} domain as a simple setup to test
    compositional behaviour and show that it is theoretically possible for a
    standard RNN to learn to behave compositionally in this domain when trained
    with standard gradient descent and provided with additional
    supervision. We then remove this additional supervision and perform a
    search over a large number of model initializations to investigate the
    proportion of RNNs that can still converge to a compositional solution. We
    discover that a small but non-negligible proportion of RNNs do reach
    partial compositional solutions even without special architectural
    constraints. This suggests that a combination of gradient descent and
    evolutionary strategies directly favouring the minority models that
    developed more compositional approaches might suffice to lead standard RNNs
    towards compositional solutions.
\end{abstract}


\section{Introduction}

The last few years have seen the re-emergence of neural networks as incredibly
effective all-purpose learning systems \citep{LeCun:etal:2015}. However, neural
networks still need to be specialized to specific tasks, with little or no
cross-task transfer, and they require huge amounts of training data to perform
well \citep{Lake:etal:2016}. One reason for these limitations is that they
are not able to perform \emph{compositional learning}, that is, to discover and
store \emph{skills} that are common across problems, and to re-combine them in a
\emph{hierarchical} fashion to solve new challenges \cite{Schmidhuber:1990}. The
ability to perform compositional learning would provide better generalization
and therefore result in a reduction of sample complexity of learning algorithms. Further down the road, compositional methods might be key ingredients in the formulation of full-fledged, general-purpose lifelong learning systems.

In stark contrast to neural networks, compositional abilities -- as it is
generally agreed -- are a core aspect of human cognition
\citep{Minsky:1986,Fodor:Pylyshyn:1988,Fodor:Lepore:2002,Lake:etal:2016}. Direct
evidence for the claim that humans are compositional learners was provided by
\citet{Schulz:etal:2016}, who explored human intuitions about functions through
extrapolation and completion experiments, and concluded that these intuitions
are best described as compositional.\footnote{Specifically, the authors show that
participants ``prefer compositional over non-compositional function
extrapolations, that samples from the human prior over functions are best
described by a compositional model, and that people perceive compositional
functions as more predictable than their non-compositional but otherwise similar
counterparts.'' \cite{Schulz:etal:2016}} Strikingly,
\citet{Piantadosi:Aslin:2016} have shown that 3.5-4.5 year olds generalize
function composition above chance even when they have not been trained on the
composition process itself.

In view of the clear advantages of compositional learning, there has been a
growing interest in equipping neural networks with compositional abilities (the
literature is partly reviewed in Section~\ref{sec:related}). As opposed to that
line of research, in this paper we explore the compositional generalization
capabilities of standard recurrent neural networks (RNNs, \citealp{Elman:1990})
without any special architectural constraints. We first introduce the
\emph{lookup table composition} domain as a simple and highly flexible setup to
test compositional behaviour (Sections~\ref{sec:lookups} and \ref{sec:seq2seq}).
We then analytically sketch how an RNN can represent and compose functions, and
demonstate that it can learn this behaviour if explicit supervision is provided
on its hidden layer (Section~\ref{exp1}). Finally, we attempt to let RNNs
discover a compositional solution to our sets of tasks via standard
example-driven gradient-descent-based training (Section~\ref{exp2}) and examine
what proportion of the trained models do discover such a solution. 

While an average training run does not
converge to a RNN that behaves compositionally, in a large random
search over initializations a (partially) compositional solution is
discovered in a small but non-negligible number of
cases. Interestingly, convergence to a compositional solution was not
determined by the initialization of the model, but rather by seemingly
minor random factors such as order of task presentations and weight
updates. All in all, our results suggest that a combination of
gradient descent and evolutionary strategies directly favouring the
minority models that developed more compositional approaches might
lead to inducing compositional RNNs without special architectural
constraints.

\section{Composing table lookup functions}
\label{sec:lookups}

The ability to discover and apply \emph{function composition} is a natural
starting point to test the compositional skills of a learning system. Mastering
function composition increases the  expressivity of a system: It is only
through function composition that a system could handle recursion, allowing it
to process an infinite number of objects through finite means. This
``combinatorial infinity'' property can be observed in language (e.g., in the
power to construct sentences of unbounded length by multiple clause embedding,
\citealp{Hauser:etal:2002}) and other cognitive domains (e.g., planning or
mathematical reasoning).

Since our focus is on composition itself, rather than on the ability of the
learning system to solve sophisticated primitive tasks, we consider a set of
such tasks that only require rote memorization, namely the \emph{table lookup}
tasks. For all possible bit strings of a fixed length, a table lookup function
is an arbitrary bijective mapping of the string set onto itself. For example,
there are 8 possible 3 bit strings, and consequently $8! = 40,320$ distinct
mapping tables, as each possible permutation of the string set corresponds to
one distinct output assignment (see for example mapping tables $g$ and $c$ in
Table~\ref{tab:composition-examples}).

As the table lookup functions share domain and co-domain, the output of any
function is a well-formed input for any other function, and thus we can generate
an infinite number of new functions by composition. For example, if
$g(000)=010$, $g(010)=001$, $c(001)=000$ and $c(010) = 101$, we can apply
function compositions such as $cg(000)=101$, $gg(000)=001$, $cgc(001)=101$, and
so forth.

In an attempt to make the presence of composition more explicit,
we require our models to produce, as output of a composed table lookup, the
output of the intermediate steps as well.\footnote{Results are however stable
if, instead, we only request models to produce the final output.} This is
illustrated for compositions of two 3-bit string tables in
Table~\ref{tab:composition-examples}.

An important advantage of the table lookup tasks is the clear separation of
atomic and composed tasks, which allows for a straightforward evaluation of
compositional behaviour on part of the learning system. Let us consider a
learning system that has mastered the mappings $g$ and $c$ and has been
presented with several input-output pairs of the composed function $cg.$ If it
behaves compositionally and if it understands that the mapping $cg$ is a
composition of the two underlying mappings $g$ and $c$, it should have no
trouble in producing the correct output for unseen inputs of this composed
mapping, i.e., it should be able to perform \emph{zero-shot} generalization.

\begin{table}
    \begin{tabular}{lcr}
        \multicolumn{3}{c}{Atomic} \\
        \multicolumn{3}{c}{$g$} \\
        \toprule
        000 &$\rightarrow$ &010 \\
        001 &$\rightarrow$ &110 \\
        010 &$\rightarrow$ &001 \\
            &$\dots$ \\
        \bottomrule
    \end{tabular}
    \hfill
    \begin{tabular}{lcr}
        \multicolumn{3}{c}{Atomic} \\
        \multicolumn{3}{c}{$c$} \\
        \toprule
        000 &$\rightarrow$ &100 \\
        001 &$\rightarrow$ &000 \\
        010 &$\rightarrow$ &101 \\
            &$\dots$ \\
        \bottomrule
    \end{tabular}
    \center
    \begin{tabular}{lcr}
        \multicolumn{3}{c}{Composed} \\
        \multicolumn{3}{c}{$cg$} \\
        \toprule
        000 &$\rightarrow$ &010101 \\
        001 &$\rightarrow$ &110111 \\
        010 &$\rightarrow$ &001000 \\
            &$\dots$ \\
        \bottomrule
    \end{tabular}
    \caption{Examples of two atomic 3-bit lookup tasks $g$ and $c$ and
      their composition $cg$. Note that the output of the composition
      includes the output of the intermediate step (in this case, of
      applying function $g$).}
    \label{tab:composition-examples}
\end{table}

\section{Compositional table lookups as sequence-to-sequence learning}
\label{sec:seq2seq}

\begin{figure*}
    \begin{subfigure}[b]{\linewidth}
        \centering
        \includegraphics[width=0.5\linewidth]{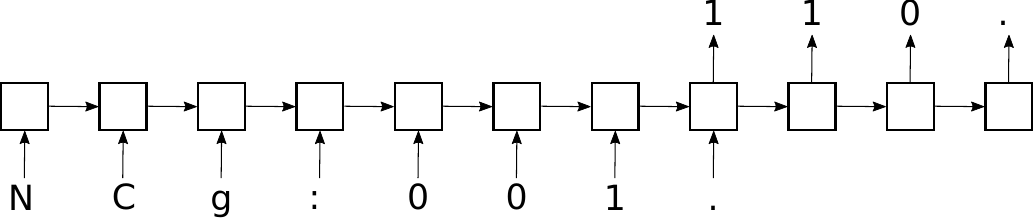}
        \caption{$g(001)$}
        \label{fig:seq2seq:a}
    \end{subfigure}
    \begin{subfigure}[b]{\linewidth}
        \centering
        \includegraphics[width=0.7\linewidth]{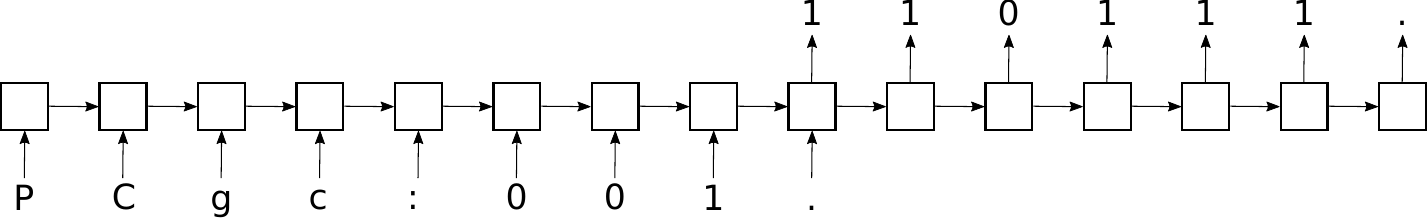}
        \caption{$cg(001)$}
        \label{fig:seq2seq:b}
    \end{subfigure}
    \caption{Compositional table lookups as sequence-to-sequence learning (refer
    to Table \ref{tab:composition-examples} for the lookup tables).}
    \label{fig:seq2seq}
\end{figure*}

We approach the lookup table tasks from the perspective of
character-level sequence-to-sequence learning
\cite{Sutskever:etal:2014}.  Figure~\ref{fig:seq2seq:a} shows an
example of a single episode in which the network reads the input
\io{NCg:001.}, which represents the atomic task $g(001)$. The network
starts producing the output, \io{110.}, as soon as the dot character
signals the end of the input string. The output is considered correct
if the output string consists of the correct outcome of function
application specified on input, and if the output string is terminated
with a dot. Similarly, Figure~\ref{fig:seq2seq:b} shows a sample
episode of function composition $cg(001)$.\footnote{Note that we
  present the function codes in order of application to the network.
  In this example, \io{gc} requires applying lookup $g$ before $c$.} 
The objective therefore bears some similarities to a
traditional language modelling objective \cite{Mikolov:etal:2010}.

\begin{figure}
    \includegraphics[width=0.9\linewidth]{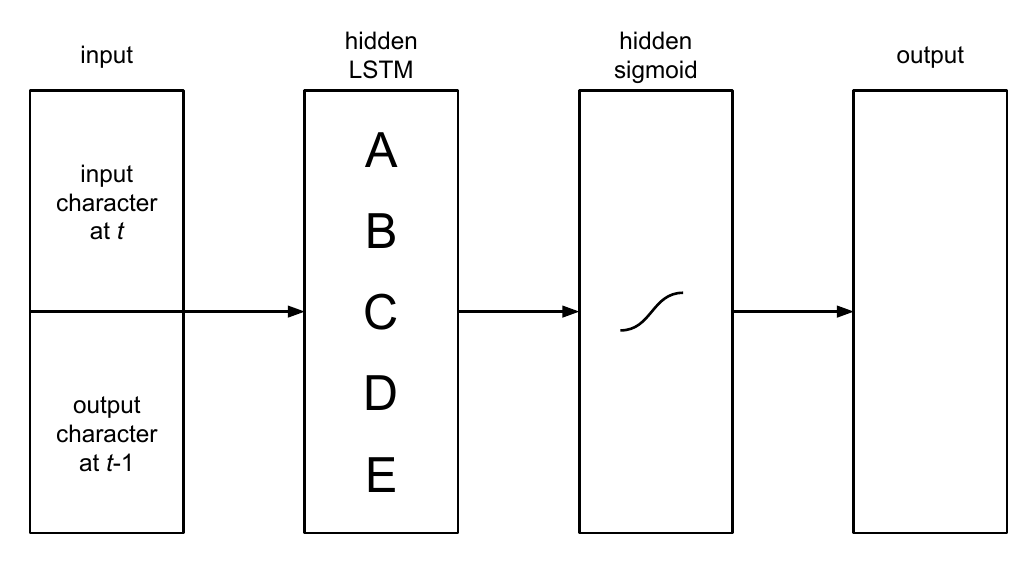}
    \caption{Model architecture. The LSTM layer is further subdivided into 5
    segments in Experiment 1; see Section~\ref{exp1} for further details on the
    segments.}
    \label{fig:model_diagram}
\end{figure}

The model used in the experiments is a neural network with two hidden layers: a
recurrent LSTM \cite{Hochreiter:Schmidhuber:1997} layer with 60 units
(restricted to 29 units for the experiment in Section~\ref{exp1}) and a sigmoid
layer with 10 units (Figure~\ref{fig:model_diagram}).  The input layer of the
model consists of concatenated one-hot vector encodings of the input character
(or a space \io{ } when the whole prompt has been read) and the output
character of the previous step. This architecture is motivated by the
compositional solution we propose in Section~\ref{exp1} below. The input
vocabulary of the network consists of characters \io{P}, \io{N}, \io{C}
specifying the type of task (atomic or composed), lookup table codes \io{a}, \io{b} through \io{h}, bits
\io{0} and \io{1}, punctuation marks \io{:} and \io{.}, and space \io{ }. The
output layer is a softmax layer with three units for the three possible output
characters \io{0}, \io{1}, and \io{.}. At each step the output character with
the highest score is selected. 
\\
\\
The model and training were implemented in PyTorch.\footnote{\url{http://pytorch.org/}} The Adam algorithm \cite{kingma2014adam}
was used for optimization. Other details of the training procedures are
described separately for the two experiments in sections \ref{exp1} and
\ref{exp2} below.

\section{Experiments}

\subsection{Experiment 1: Can an RNN encode a compositional solution through a
finite-state automaton?}
\label{exp1}

In the first experiment, we tested whether there exist such weights of a
character-level RNN that give place to a model demonstrating compositional
behaviour. If this were the case, it would show that in principle a RNN of the
described architecture can achieve strong generalization capabilities and
produce correct outputs for unseen inputs of known composed functions. 

If we limit ourselves to a specific maximum number of lookup tables of
finite length bit strings and a maximum number of possible composition
steps, a simple approach to model lookup table compositions is through
finite-state automata (FSA).\footnote{FSAs do not support infinite
  recursion, but they can handle arbitrarily deep embeddings, although
  at some considerable computational cost.}  Instead of producing the
weights directly by hand, we designed an encoding scheme for the
recurrent layer that represents the state of such an automaton and
we directly supervised the output of this layer to conform to this
scheme. Once this encoding scheme has been learned, we proceeded to
training the mapping from state representations to output characters.

The state of the FSA solving lookup table compositions needs to encode the
following pieces of information: a (finite) stack of atomic tasks to perform,
the input bit string for the current task, an index into the output string, and
the bits produced so far for the current task (as they form the input string of
the following task). Such state information can be represented by the recurrent
layer in the form of a binary code. Specifically in our case, the units of the
hidden layer are divided into the following segments
(Figure~\ref{fig:model_diagram}):
\begin{itemize} 
    \item segment \seg{A} (8 units) encodes the current atomic task (out of 8)
        using one-hot encoding,
    \item segment \seg{B} (8 units) encodes the following atomic task (if any)
        using one-hot encoding;
        segments \seg{A} and \seg{B} represent the "call" stack,
    \item segment \seg{C} (6 units) encodes the input bit string using three
        one-hot vectors of size 2 units,
    \item segment \seg{D} (3 units) represents an index into the output string,
        i.e., it encodes which bit (index) of the output string should be output
        (none, first, second, or third),
    \item segment \seg{E} (4 units) stores the characters output on previous steps 
        of the current mapping task (two one-hot vectors, each of size 2).
\end{itemize} 
This encoding consists of a total of 29 units. Note that the hidden layer only
records the information provided by the input string and does not encode the
actual characters to output; in particular, it does not remember the lookup
tables, as this is left to a separate sigmoid unit layer and to
the output layer. The units of the sigmoid layer have non-zero weights only on
connections from segments \seg{A} (task), \seg{C} (input), and \seg{D} (output
index), as these are the only segments directly affecting which character should
be output at each step.

As an example, let us consider the input \io{PCgc:001.}. All units of the
recurrent layer output values of approximately zero at the beginning of the
episode. As the network reads the input, appropriate units change their output
from zero to values close to one. Specifically, after reading the third
character \io{g}, the seventh unit of segment \seg{A} starts to output value 1,
as $g$ is the first mapping to be performed. After reading the next character,
\io{c}, the third unit of segment \seg{B} activates, since the second mapping to
perform is $c$. Next, the input string is encoded in segment \seg{C}. Finally,
as the network reads the dot character, the first unit of segment \seg{D}
activates, signaling that the network should produce the first output bit of the
first mapping $g(001).$ On the next step, this first output bit is
``stored''\footnote{These actions are performed by means of transitions in the
recurrent layer rather than hand-coded copy-and-paste operations.} in segment
\seg{E} and the second unit of segment \seg{D} is activated, signaling that the
second output bit of the mapping $g(001)$ should be produced. Once all output
bits of the mapping $g(001)$ are generated, the contents of segment \seg{B} are
``moved'' to \seg{A} (i.e., the task $g$ is removed from the top of the call
stack), the contents of \seg{E} (representing the bit string \io{11}) are
``moved'' to the first four units of segment \seg{C}, and the output character
produced on the previous step (\io{0}) is ``appended'' to segment \seg{C}.
Furthermore, the first unit of segment \seg{D} is activated, indicating that the
first output bit of the mapping $c(110)$ should be produced.

\paragraph{Training and evaluation} The weights of this network are trained in
two phases, as the encoding scheme needs to be learned prior to the mapping of
state representations to output characters. In the first phase, the network
learns to set the right transitions between recurrent layer states in the
presence of specific input. For this, we generated random pairs of input-output
strings of the correct form for both atomic and composed tasks. For each
input-output pair (or \emph{episode}) we furthermore generated a sequence of
binary vectors of length 29 representing the target values of the recurrent
layer at each time step using the encoding scheme described above. We used these
vectors as direct supervision on the output of the recurrent layer (using
mean-squared error loss) and trained the weights of input-to-hidden and
recurrent connections with stochastic gradient descent and backpropagation
through time, updating the weights after each episode for a total of 1,000k
training episodes. In the second phase, we proceeded to train the mappings
from recurrent layer state representations to output characters (using
cross-entropy loss). In this step, we froze the weights of input-to-hidden and
recurrent connections and only trained the connections between the recurrent and
sigmoid layers and from the sigmoid to the output layer. We generated a random
task set consisting of 8 atomic 3-bit tasks. We sampled 1,000k episodes from the
atomic tasks, and trained the network using stochastic gradient descent,
updating the weights after each episode.  Note that there is no need for
training on composed tasks as the ability to produce correct output strings for
composed tasks should follow automatically once the atomic tasks are learned,
thanks to (1) the specific pre-training of recurrent layer transitions in the
first phase of training, and (2) the connections between the two hidden layers.
The final network was evaluated on all possible inputs to the 8 atomic and the
64 associated composed tasks. 

\paragraph{Results} The trained network produces correct output across both
atomic and composed tasks in 96\% of the cases. Note that the network has not seen any
composed tasks during training and its generalization capabitilies are purely
due to the transition logic of the recurrent layer and the connectivity between
the two hidden layers. This experimental result confirms that the weight-space
of a RNN can implement the specific compositional FSA solution that we sketched
above, and can learn it when provided with direct supervision on the structure
of the automaton. However, the results do not imply that the devised binary
encoding is a natural solution to the problem, nor that a RNN can in practice
discover a compositional solution when provided with input/output examples as the only training signal. These questions are pursued in the following section.

\subsection{Experiment 2: Search for a compositional RNN over model
initializations}
\label{exp2}

To investigate the proportion of RNNs that converge to a compositional solution
when training on input/output examples only, we run a
large random search over model initializations, train the networks with standard
cross-entropy loss and gradient descent on a set of atomic and composed tasks,
and then test them on zero-shot compositional generalization, to check if any of
the RNNs in the batch discovered a compositional solution, this time without any
external guidance.\footnote{We also explored a pure search-based approach where
networks are randomly initialized (in the same range specified below) and
directly tested without further training. No network of this sort behaves better
than chance level.}

\begin{figure}
    \includegraphics[width=0.95\linewidth]{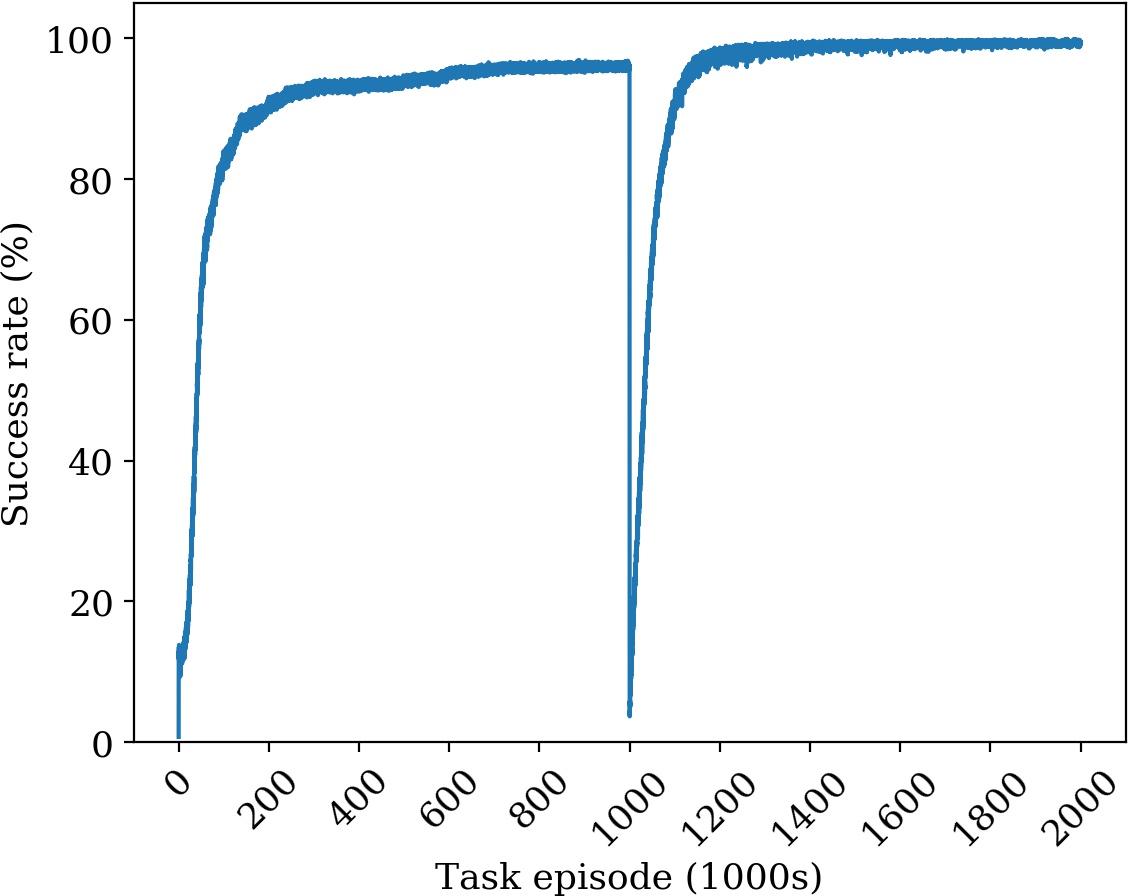}
    \caption{Average success rate (percentage of correct outputs in the most
    recent 100 episodes) during atomic (first 1,000k episodes) and atomic+composed lookup training (second 1,000k episodes). For this
    figure, we generated five random task sets of 8 atomic and 64 composed
    tasks, and trained 10 randomly initialized networks on each task set. The
    success rate shown is averaged across these $5 \times
    10$ networks.}
    \label{fig:success_rates_lookups}
\end{figure}

\paragraph{Training} We generated a random task set of 8 atomic 3-bit
lookup tasks and the corresponding 64 pairwise compositions. 50k
models with the same architecture as in the previous experiment were
trained in two phases, at first with episodes drawn from atomic tasks
only (1,000k training episodes) and later with tasks sampled across
both atomic and composed tasks (further 1,000k episodes). For each
composed task, we withheld two input strings that were used for
evaluation (see below).  Each model was initialized with random
weights drawn from a uniform distribution $\mathcal{U}(-0.1, 0.1)$ and
with zero biases, and was trained with backpropagation through
time. We used the cross-entropy loss and updated the weights after
each episode (stochastic gradient descent). The training of each model
was performed asynchronously on 40 CPUs in parallel. As shown in
Figure~\ref{fig:success_rates_lookups}, by the end of the first phase
most of the models mastered all atomic tasks and, by the end of the
second, they further mastered all composed tasks when fed seen inputs.

\paragraph{Evaluation} We evaluate the compositionality of trained models by
their zero-shot generalization to withheld inputs on composed tasks. In our
case, we test the models on $2 \times 64 = 128$ unseen composed task+input
combinations, and report the percentage of correctly answered test items as
\emph{generalization performance}.

\paragraph{Baselines} We evaluated several random baselines. The
simplest one, \bas{random-output}, produces output strings by
randomly sampling from the set of the three possible output characters \io{0},
\io{1}, and \io{.}. As learning the expected form of the output sequence (three
bits and a dot character for atomic tasks, six bits and a dot for
composed tasks) is relatively easy for a network, we also evaluated baseline
\bas{random-wellformed-output}, which randomly samples six bits and appends the
dot for all test items. Lastly, we evaluated baseline
\bas{random-task-code}, which is equivalent to the model employed in the
experiment, but whose training input strings for composed tasks (such as
\io{PCcf:010.}) were altered so that there is no consistent relation between the
task codes (\io{cf}) and the underlying tasks. This last baseline is meant to
capture any kind of statistical biases in the task set (such as shared
input-output mapping pairs for a subset of inputs across two lookup tables) that
would not be captured by a fully random baseline.  Baselines \bas{random-output}
and \bas{random-wellformed-output} were evaluated 10k times on the test dataset,
while baseline \bas{random-task-codes} was evaluated for 1k trained models.

\begin{figure}
    \includegraphics[width=0.95\linewidth]{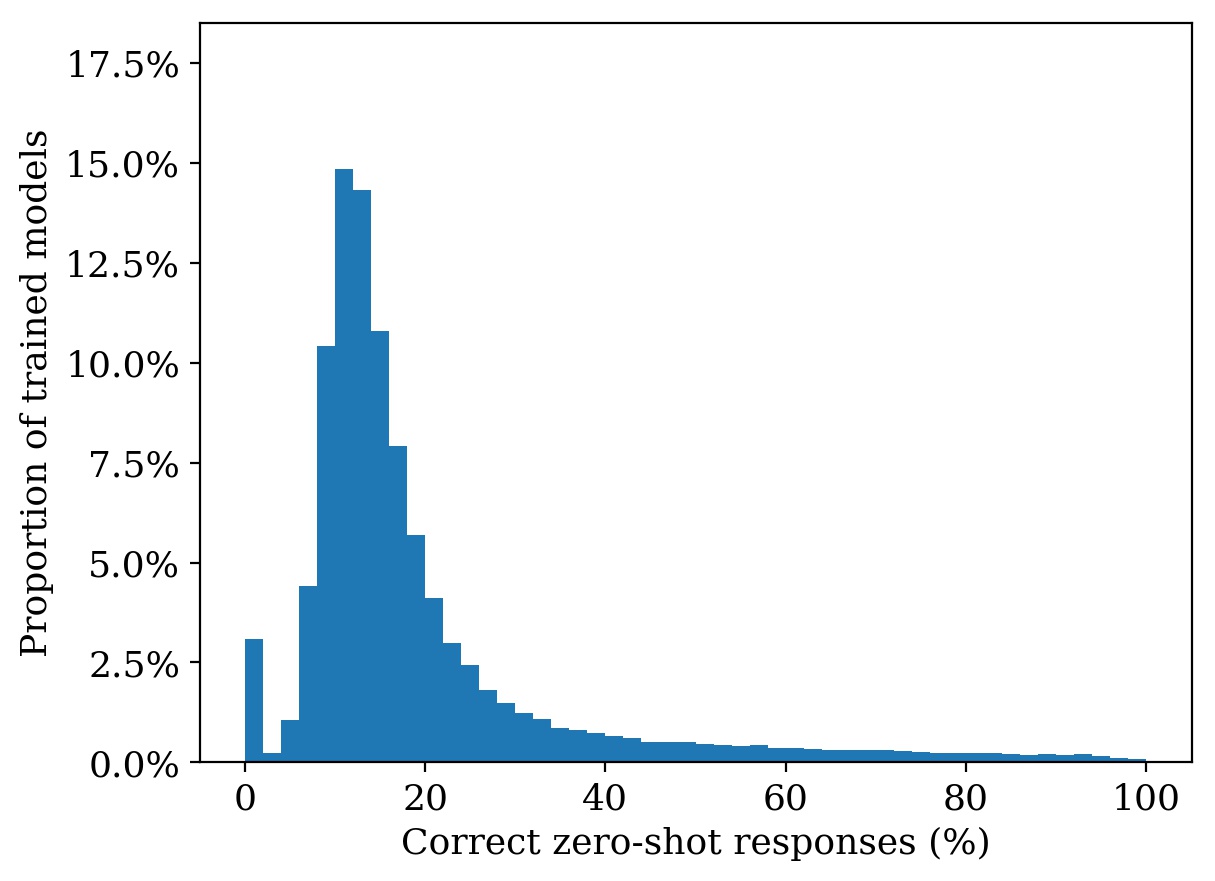}
    \caption{Generalization performance of the 50k models trained in the search
    experiment.}
  \label{fig:random_search_results}
\end{figure}

\begin{figure}
    \includegraphics[width=0.95\linewidth]{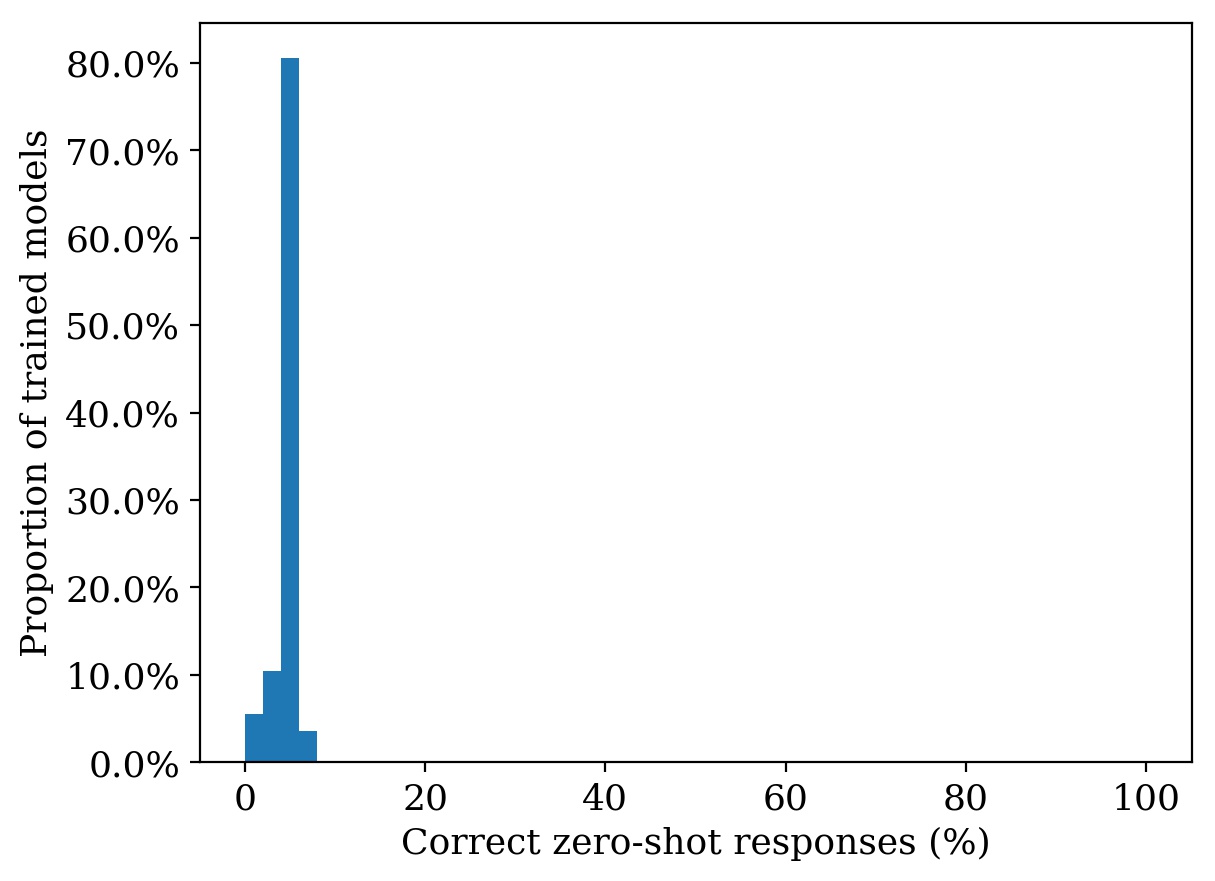}
    \caption{Generalization performance of the 1K models of baseline \bas{random-task-codes}.}
    \label{fig:german_baseline}
\end{figure}

\paragraph{Results} Figure~\ref{fig:random_search_results} shows the overall
distribution of the 50k runs in terms of generalization performance. For
comparison, Figure~\ref{fig:german_baseline} shows the baseline
\bas{random-task-codes} generalization performance distribution across 1k runs
of models, and Table~\ref{tab:performance} shows average generalization
performance for all baselines. Most runs in
Figure~\ref{fig:random_search_results} show performance well above the
baselines, but they are far from successful at generalization.  However, we also
observe a tail of models that do generalize very well: $\approx{}2\%$ models
reach zero-shot accuracy $>80\%$, and $0.75\%$ models reach zero-shot accuracy
 $>90\%$, while no baseline model ever achieves performance anywhere close
to these levels. We thus conclude that RNNs trained with standard gradient
descent methods on a task involving composition can, occasionally, discover a
compositional solution that allows them to generalize zero-shot. The chances to randomly
stumble upon such a RNN are, however, quite slim.

\begin{table}
    \centering
    \begin{tabular}{l r}
        \toprule
        &Generalization \\
        &performance (\%) \\
        \midrule
        RNN &19.60 \\
        \midrule
        Baselines & \\
        - \bas{random-output} &0.00 \\
        - \bas{random-wellformed-output} &0.01 \\
        - \bas{random-task-codes} &4.56 \\
        \bottomrule
    \end{tabular}
    \caption{Average generalization performance in the random RNN search and for all three
    baselines.}
    \label{tab:performance}
\end{table}

\begin{figure}
    \includegraphics[width=0.95\linewidth]{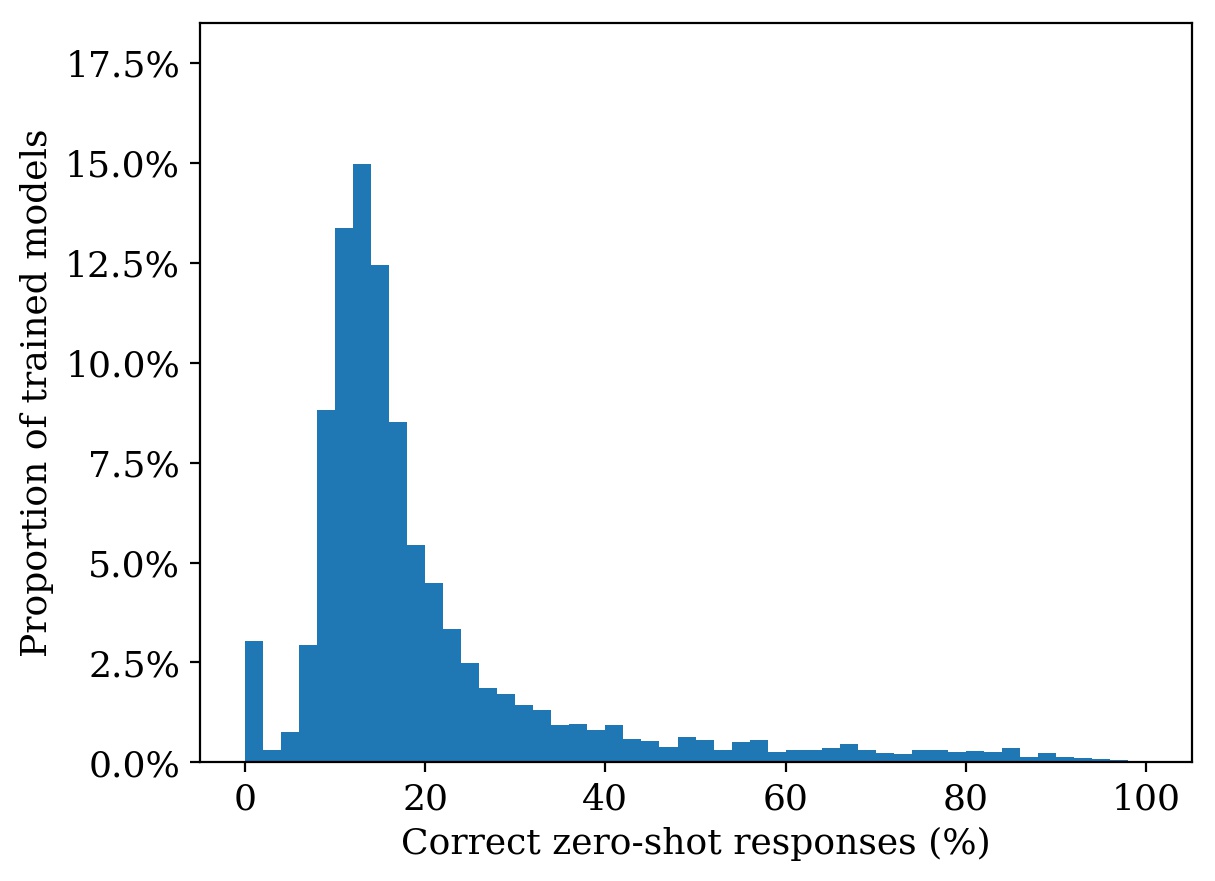}
    \caption{Distribution of generalization performance of 5k models trained
    with consistent but shuffled task prompts for composed tasks.}
    \label{fig:noninformative_prompts}
\end{figure}

The first follow-up question we ask is whether the compositional RNNs learned to
parse the ``language'' of the prompts, and thus interpret, say, the \io{gc}
sequence as an instruction to apply lookup table $g$ followed by lookup table
$c$. This is a stronger form of compositionality, akin to the one we encounter
in natural language, where string composition mirrors meaning composition
\citep{Montague:1970a}.  Figure~\ref{fig:noninformative_prompts} suggests that
obfuscating the prompts so that it is no longer possible to identify the atomic
tasks involved in a compositional operation (e.g., \io{db}
consistently cues the composition of tasks $a$ and $h$) does not affect the
overall performance curve. Thus, even for the RNNs that converged to a
compositional solution, the latter is associated to arbitrary codes that must be
memorized, rather than to a decompositional analysis of the
prompts.\footnote{This observation is supported by additional experiments in
which the network was trained with a subset of composed tasks and tested on
unseen composed tasks. The network did not generalize well to these tasks, which confirms it is not learning to decode prompt structure (data not shown).}

\begin{figure}
  \includegraphics[width=0.95\linewidth]{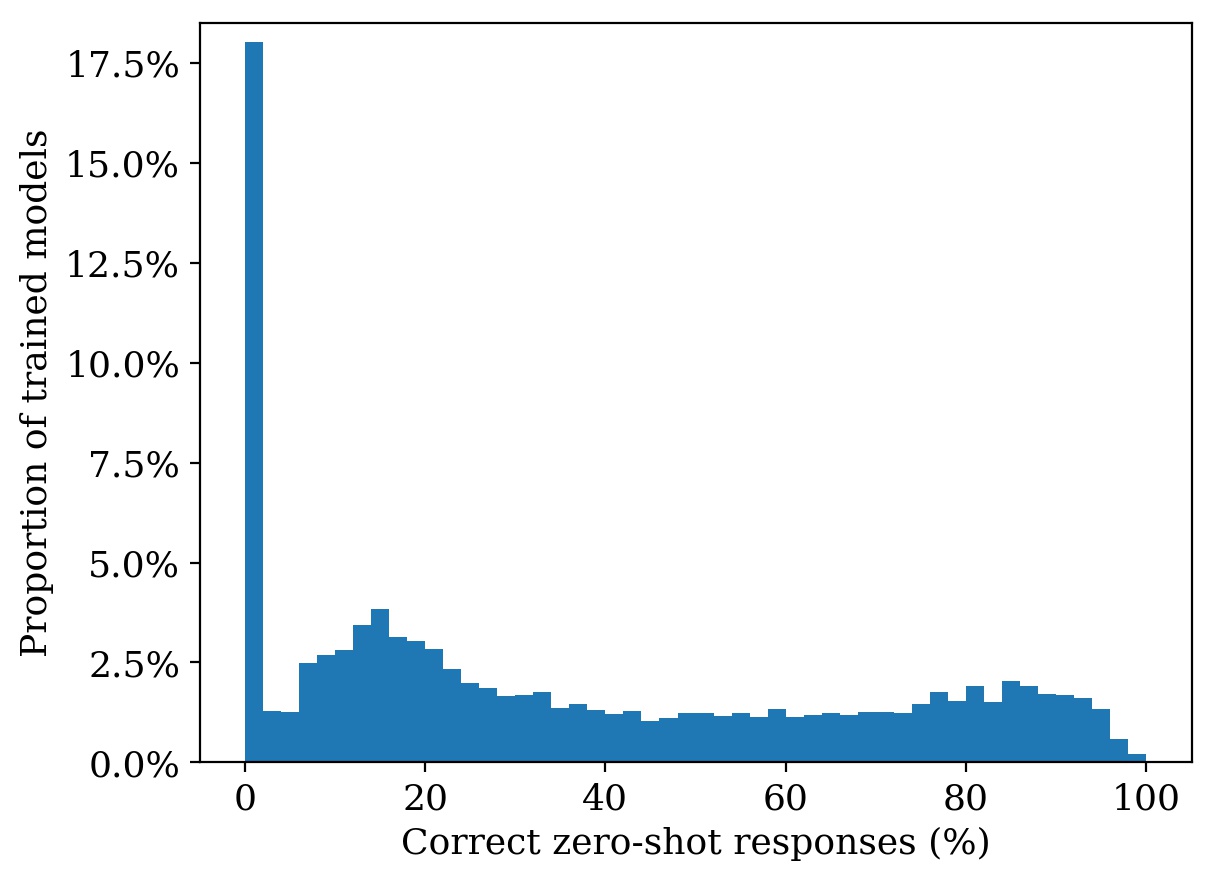}
    \caption{Distribution of generalization performance of 7k models trained
    with composed tasks only.}
  \label{fig:no_atomic_in_training}
\end{figure}

Next, we inquire about the importance of the curriculum we used in the main
experiment, where we started by teaching the model how to perform atomic
lookups, and later added compositional tasks.
Figure~\ref{fig:no_atomic_in_training} suggests that, when training on composed
tasks only, a larger number of models drop to baseline level on generalization,
but on the other hand there is also a considerably larger proportion of models
that learn to generalize correctly ($5.55\%$ of 7k trained models with zero-shot
accuracy $>90$\%). Together with the previous experiment, this suggests that it
is not only the case that successful models fail to relate atomic and composed
task codes. They are also most probably failing to exploit their knowledge of
atomic tasks when solving composed tasks. Success at zero-shot generalization of
models trained solely on composed tasks suggests that the models are inducing
their own representation of the atomic lookups while learning the composed tasks,
rather than exploiting the representations acquired from atomic tasks training.

\begin{figure}
    \includegraphics[width=0.95\linewidth]{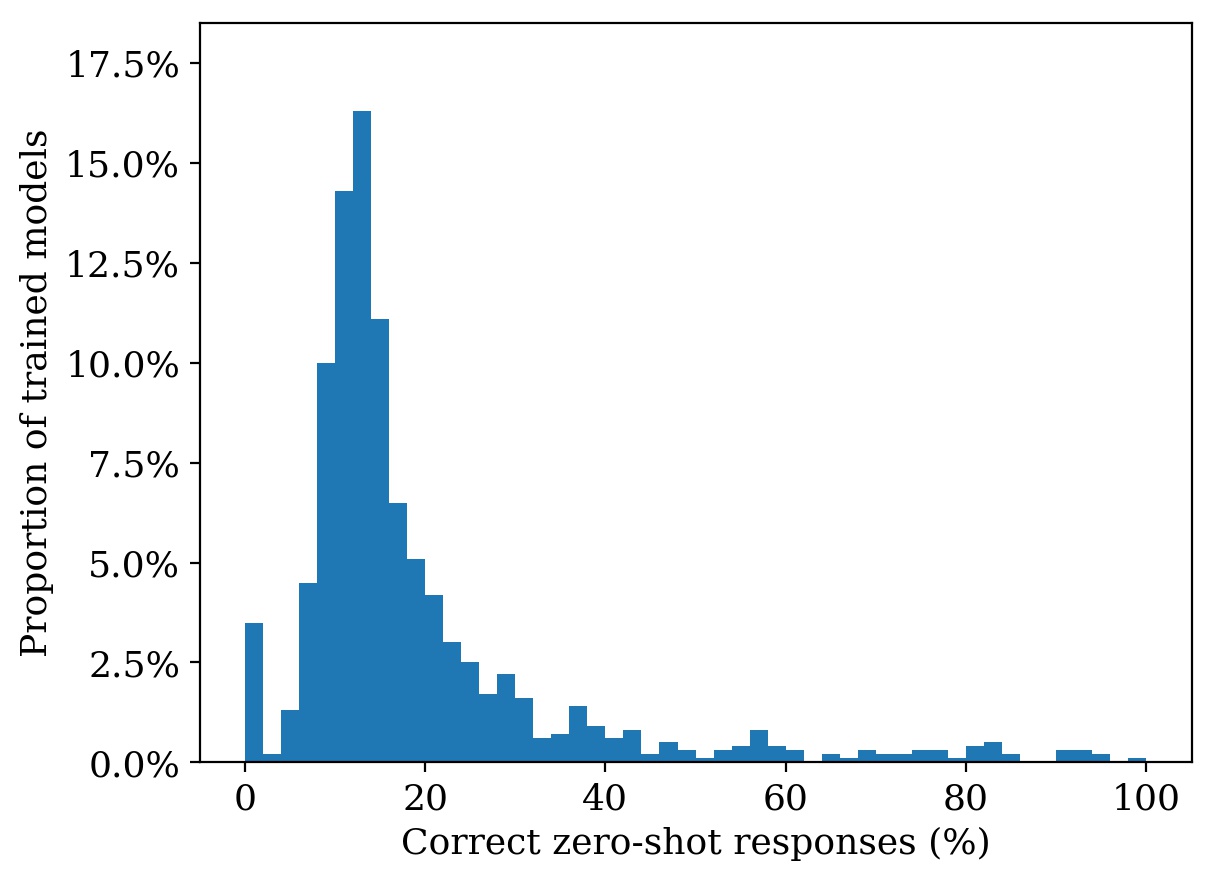}
    \caption{Distribution of generalization performance of 1k re-runs with an
    initialization that originally led to perfect generalization.}
    \label{fig:repeated_runs_successful_init}
\end{figure}

\begin{figure}
    \includegraphics[width=0.95\linewidth]{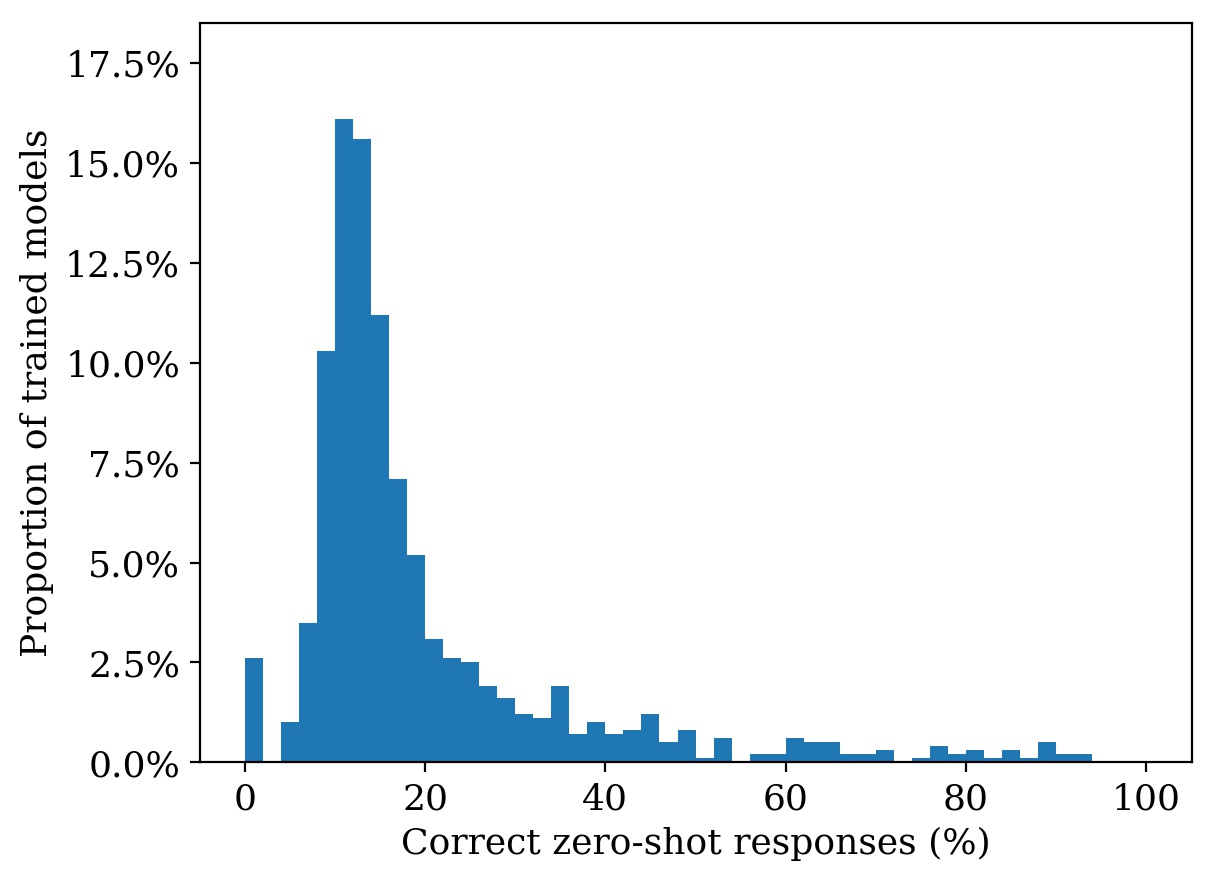}
    \caption{Distribution of generalization performance of 1k re-runs with an
    initialization that originally did not generalize to unseen inputs of
    composed tasks.}
    \label{fig:repeated_runs_unsuccessful_init}
\end{figure}

Finally, we test whether it is the different
initializations and their properties that lead some models to generalize better
than others. In Figure~\ref{fig:repeated_runs_successful_init}, we report the
distribution of 1k re-runs with one of the most successful initializations in
the random search experiment. In
Figure~\ref{fig:repeated_runs_unsuccessful_init}, we report the same
distribution for one of the worst initializations in the original experiment.
Surprisingly, the figures suggest that initializations have \emph{no} effect on
the odds to converge to a successful model. Evidently, the determining factor is
the (random) order in which tasks are presented and weights updated during
training.

\section{Related work}
\label{sec:related}

The idea of statistical learning systems, and more specifically neural networks
capable of skill composition has been around for a long time, particularly in
the domain of reinforcement learning, where it is natural to frame higher-level
tasks as hierarchical compositions of simpler actions
\citep{Schmidhuber:1990,Sutton:etal:1999,Barto:Mahadevan:2003,Taylor:Stone:2009}.
In this domain, composition almost always consists in temporally concatenating
sequences of actions, and thus lacks the recursive properties of proper function
composition discussed in Section~\ref{sec:lookups}.

As early as \citet{Singh:1992}, a standard approach to neural-network-based
composition has been to structure the network into a set of modules that are
trained to solve specific tasks, plus a controller or gating system that learns
which module to call at each point in time. The modular approach has recently
been greatly extended, and applied to problems, such as visual question
answering, that require proper function composition
\cite{Andreas:etal:2016b,Andreas:etal:2016,Hu:etal:2017,Johnson:etal:2017}.
While the tasks tackled by these models are much more complex than table lookup
composition, the models themselves must make strong \emph{a priori} assumptions
about the structure of the controller, the set of modules and how they can be
combined. They moreover require direct supervision on the module sequence to be
applied, or some degree of hand-coding of module functionality. For these
reasons, it is difficult to see how such approaches could scale up to genuine
lifelong learning scenarios, where one is faced with an open-ended set of new
skills to be acquired.

A very promising recent work \cite{Sahni:etal:2017} focuses on skill
composition. In the proposed architecture, separate skill networks produce
embeddings that are then (possibly recursively) composed by a differentiable
composition function. Still, the system requires separate training of the skill
networks and composition function.

Finally, compositional skills of sequence-to-sequence recurrent networks have
been recently evaluated in the framework of a simple compositional navigation
environment, showing that RNNs fail when generalization requires systematic
compositional skills \cite{Lake:Baroni:2017}.

\section{Discussion}

We have studied the question of whether a recurrent neural network~(RNN) can
learn to solve a function composition task compositionally, that is, by storing
the constituent functions, and combining them to solve new problems in a
zero-shot fashion.

In the specific table-lookup domain we considered, we find that it is
theoretically possible for a RNN to learn to behave compositionally in the sense
above, at least up to a finite number of compositions. Moreover, a large random
search shows that a certain proportion of RNNs converged to a compositional
solution as indicated by their successful generalization to unseen inputs of
composed tasks that is well above chance levels. These seem, however, to perform
a weaker form of composition that does not rely on analyzing the composed task
prompts, suggesting that networks represent the latter as single undecomposable
units that index specific atomic or composed tasks.

Our results show that initializations, at least in the range
explored in our experiments, have very little effect on the final performance of
the networks, suggesting instead that seemingly minor random factors such as
order of task presentations and weight updates determine whether the path taken
by the model is memorization-based or compositional.

In future research, we would like first of all to gain a better understanding of
the compositional strategies induced by the best models. One approach to do this
is through extracting FSAs from learned RNNs
\cite{giles1992learning,Weiss:etal:2017}, i.e., the opposite of the
FSA-into-RNN process that has been carried out in Section~\ref{exp1}. 

Second, we would like to devise new training regimes leading RNNs to fully
compositional solutions in more stable ways. One key insight here is that our
current training regime does not explicitly reward zero-shot generalization,
which is only evaluated at test time for models that have been trained with
standard cross-entropy-based gradient descent on a large number of repetitive
examples. Gradient-based techniques are hard to apply to a generalization
objective, which cannot be naturally formulated in differentiable terms. Thus,
in our following experiments we plan to switch to more flexible evolutionary
techniques, using zero-shot generalization on held-out compositions as our
fitness criterion. Switching to an evolutionary approach also opens up
interesting possibilities in terms of neural network plasticity, and we would
like to explore architectures that grow larger during training, as they might
encourage more modular structures that in turn should favour compositionality
\cite{Soltoggio:etal:2017}. At the same time, with the huge power afforded by
these methods come more difficulties in making them converge. We informally
experimented with the popular NEAT algorithm \cite{Stanley:Miikkulainen:2002}
applied to our lookup tables, but we were not able to get it to solve even the
atomic tasks.

Third, future work should take advantage of the full potentials of the table
lookup domain. These include testing the generalization to more compositions,
working with longer bit strings, and testing the comprehension of the prompt
``language`` by evaluating on zero-shot compositions instead of zero-shot inputs
only. Finally, we would like to explore to what extent results obtained in this
domain generalize to other compositional problems (for example, in language).

\section*{Acknowledgments}
We thank Allan Jabri and Rahma Chaabouni for earlier versions of the code we
used in our experiments. We also thank Angeliki Lazaridou, Tomas Mikolov,
Brenden Lake, David Lopez-Paz, Jos\'{e} Hern\'{a}ndez-Orallo, Klemen Simonic,
Armand Joulin and Alexander Miller for inspiration, ideas and feedback.

\bibliography{marco,additional}
\bibliographystyle{icml2018}
\end{document}